# PREDICTION-BASED GNSS SPOOFING ATTACK DETECTION FOR AUTONOMOUS VEHICLES

**Sagar Dasgupta**\*
**Ph.D. Student**
Department of Civil, Construction and Environmental Engineering
University of Alabama, Tuscaloosa
Tel: (864) 624-6210; Email: sdasgupta@crimson.ua.edu

**Mizanur Rahman, Ph.D.**
**Postdoctoral Fellow**
Center for Connected Multimodal Mobility ($C^2M^2$)
Glenn Department of Civil Engineering, Clemson University
127 Lowry Hall, Clemson, SC 29634
Tel: (864) 650-2926; Email: mdr@clemson.edu

**Mhafuzul Islam**
**Ph.D. Student**
Glenn Department of Civil Engineering, Clemson University
351 Flour Daniel, Clemson, SC 29634
Tel: (864) 986-5446; Fax: (864) 656-2670
Email: mdmhafi@clemson.edu

**Mashrur Chowdhury, Ph.D., P.E., F.ASCE**
**Eugene Douglas Mays Professor of Transportation**
Glenn Department of Civil Engineering, Clemson University
216 Lowry Hall, Clemson, SC 29634
Tel: (864) 656-3313   Fax: (864) 656-2670
Email: mac@clemson.edu

\*Corresponding author

Word count:  4,898 words text + 4 table x 250 words (each) = 5,898 words

Paper accepted for presentation at the Transportation Research Board 100th Annual Meeting.

**ABSTRACT**
Global Navigation Satellite System (GNSS) provides Positioning, Navigation, and Timing (PNT) services for autonomous vehicles (AVs) using satellites and radio communications. Due to the lack of encryption, open-access of the coarse acquisition (C/A) codes, and low strength of the signal, GNSS is vulnerable to spoofing attacks compromising the navigational capability of the AV. A spoofed attack is difficult to detect as a spoofer (attacker who performs spoofing attack) can mimic the GNSS signal and transmit inaccurate location coordinates to an AV. In this study, we have developed a prediction-based spoofing attack detection strategy using the long short-term memory (LSTM) model, a recurrent neural network model. The LSTM model is used to predict the distance traveled between two consecutive locations of an autonomous vehicle. In order to develop the LSTM prediction model, we have used a publicly available real-world comma2k19 driving dataset. The training dataset contains different features (i.e., acceleration, steering wheel angle, speed, and distance traveled between two consecutive locations) extracted from the controlled area network (CAN), GNSS, and inertial measurement unit (IMU) sensors of AVs. Based on the predicted distance traveled between the current location and the immediate future location of an autonomous vehicle, a threshold value is established using the positioning error of the GNSS device and prediction error (i.e., maximum absolute error) related to distance traveled between the current location and the immediate future location. Our analysis revealed that the prediction-based spoofed attack detection strategy can successfully detect the attack in real-time.

**INTRODUCTION**

Automated vehicles require reliable Positioning, Navigation, and Timing (PNT) services in order to perform their autonomy functions (*1*). For example, autonomous vehicles (AVs) use PNT services to localize and navigate themselves along the roadway (*2*). Safety and operation of automated vehicles can also be significantly affected by an unreliable PNT service as AVs rely on PNT services to navigate themselves through desired routes to a destination (*3*). Therefore, reliable PNT services are necessary for the safety and efficient operation of AVs.

PNT services provided by Global Positioning System (GPS)/ Global Navigation Satellite System (GNSS) depend on satellites and radio communications, which are subject to various threats and vulnerabilities (*2*). The most common impacts are the occlusions caused by the environments, such as high rise buildings in urban areas, walls and ceilings in garages and tunnels, and even thick clouds in the sky (*4*). Besides these natural environmental impacts, GPS/GNSS systems are also subject to some other threats, such as radio interferences (*5*), spoofing on communications (*6*) data manipulations on transmitted messages (*7*), and even disruption to the GPS/GNSS infrastructures (*7*). All these threats affect the reliability of PNT services and consequently generate significant impacts on the safe and reliable operations of AVs.

Due to lack of encryption, open-access of the coarse acquisition (C/A) codes and low strength of the signal, GNSS is vulnerable to both natural and fabricated attacks compromising the navigational capability of the vehicle (*8*)(*9*)(*10*)(*11*)(*12*)(*13*). Among them, a spoofing attack is the most sophisticated type of attack that is difficult to detect as an attacker can mimic the GNSS signal and transmit inaccurate coordinates to an AV. This will enable an attacker to change the route of an AV, leading the AV to an unsafe location, and even cause a severe traffic incident. Spoofing attacks can be categorized into three types; simplistic, intermediate, and sophisticated (*14*)(*15*)(*16*). The simplistic attack, which is easy to detect because of the absence of synchronization between the actual GNSS signal and the spoofer signal, can be created using a commercial GNSS signal simulator. For an intermediate attack, spoofer uses a portable receiver-spoofer to sync the fake signal with the actual satellite time, and adjust the Doppler frequencies, code phases of all satellite signals. As a result, it will be difficult to detect a spurious signal during an intermediate attack. As an AV uses multiple GNSS receiver antennas to continuously cross-validating the signal, a sophisticated spoofed attack can be achieved using multiple phase-locked spoofers to spoof all the antennas. Thus, it is a challenge to develop a detection model against a sophisticated spoofing attack.

In this study, we have developed a prediction-based sophisticated spoofing attack detection strategy using a recurrent neural network model, which is a long short-term memory (LSTM) model. LSTM model is used to predict the distance traveled between two consecutive locations of an autonomous vehicle. Among different recurrent neural network (RNN) structures, LSTM is the most popular one because of its capability to account for short and long-term dependencies in the time-series data.

**CONTRIBUTION OF THIS STUDY**

The contribution of our study is the development of a prediction-based spoofing attack detection strategy using an artificial recurrent neural network model, i.e., the long short-term memory (LSTM) model. The LSTM model predicts the distance traveled by an AV between two consecutive locations based on current data from AV sensors. Based on the predicted distance traveled by an autonomous vehicle between a current location and the immediate future location, an error threshold is established using the positioning error of the GNSS device and prediction error (i.e., the maximum absolute error between the ground truth and predicted distance traveled). According to our knowledge, there is no study that uses the predicted location of an AV to detect a GNSS spoofing attack using current AV sensor data.

**RELATED WORK**
We have reviewed literature related to GNSS spoofing attack models and spoofing attack detection models, as presented in the following subsections.

**GNSS Spoofing Attack**
Spoofing attacks can be divided into a four-step process: (i) deployment of transmitters; (ii) take-over strategy; (iii) control strategy; and (iv) application (*17*). A variety of configurations of the spoofer transmitter are used to achieve the desired spoofing. Single or multiple transmitters can be used for spoofing (*17*). Furthermore, the spoofer can take different strategies, e.g., (i) replication of GPS/GNSS signal polarization; (ii) Doppler frequency range ; (iii) pseudo-random noise (PRN) sequences; (iv) signal bandwidth; (v) carrier frequency; vi) reception power; and vii) modulation type, so that the attacked vehicle can't detect any anomaly (*17*). Spoofer can further match the frame structure of the data bits to make the attacked vehicle trust the spoofed signal. By changing the signal time-offsets, the spoofer can change the satellite's pseudo-ranges and achieve control over the attacked vehicle. In (*18*), the authors investigated the attack generation limitations as well as possible ways to compromise an AV navigation system on a roadway. For example, navigation attacks on an AV can be created through random GPS manipulation, which can result in inaccurate navigation instructions, such as turn right in the middle of a roadway. The authors in (*18*) designed an attack algorithm to manipulate autonomous vehicle navigation systems by generating spoofed GPS data. The purpose of manipulating the navigation systems was to generate turn-by-turn spoofed data so that the VV can be guided to a wrong destination. Authors in (*19*) evaluated the security guarantees of Inertial Navigation Systems (INS)-aided GPS tracking and roadway navigation. They demonstrated an integrated and dynamically changing GPS/INS spoofing attack, which guided AV to a wrong destination. Authors in (*20*) demonstrated a new cache side-channel attack against location privacy using an adaptive Monte-Carlo localization (AMCL) algorithm.

**Spoofing Attack Detection Models**
Typically, existing countermeasures against spoofing attacks can be generally be categorized into two classes: cryptographic techniques and anomaly detection at the signal level. In general, satellite signals are authenticated using additional signals using cryptographic techniques to make the signal undecryptable for the users who do not have the security key (*19*). Signal level detection is performed by identifying discrepancies in the physical signal waveform or arrival angle from the source of the signal. Another detection approach is to verify the signal using multiple receivers of GNSS signals to detect any anomaly (*20*)(*21*). However, spoofers can spoof such receivers by generating synced signals for multiple satellites. GNSS signal frequency drift, signal strength, and the correlation between free-running crystal oscillator and GNSS signals can also be used to detect spoofing attacks (*22*). Humphreys et al. developed a software-defined civil GPS receiver-spoofer, which can generate an attack. They also developed two software-defined user-equipment based spoofing defense methods: (i) Data Bit Latency Defense and (ii) Vestigial Signal Defense (*23*). They found that their spoofing defense methods were able to successfully detect the attack, which was created by their software-defined civil GPS receiver-spoofer. Although most of the detection techniques analyze the GNSS signal, there are few studies that focused on spoofing attack detection using GPS coordinates data. Panice et al. developed a Support Vector Machine (SVM) based spoofing detection method by comparing the estimating the current state and actual location of the unmanned aerial vehicles (UAVs) (*24*). This method is more efficient in detecting spoofing attacks where the UAV is spoofed to land than detecting attacks when UAV is attacked due to the strapdown system. In (*25*), Wang et al. developed an approach to reconstruct a lost GPS where authors used real-time driving data and GPS coordinates. Their approach reconstructs the driving route using the driving data and flag an attack if the input GPS coordinates deviate more than an error threshold from the predicted GPS coordinates (*25*).

**DATASET DESCRIPTION AND DATA PROCESSING**
In this study, we have used the real-word driving dataset from the Comma.ai, named Comma2k19 (*26*), which contains various AV sensor data. The vehicle was driven for 33 hours for 2019 segments on 280 highway in California (see **Figure 1**). Each roadway segment was 1 minute long in terms of travel time. All the roadway segments are located between San Jose and San Francisco, and the total length is 20km. The AV used by the Comma.ai has a front-facing camera, thermometers, and 9-axis Inertia Measurement Unit (IMU) for collecting data.

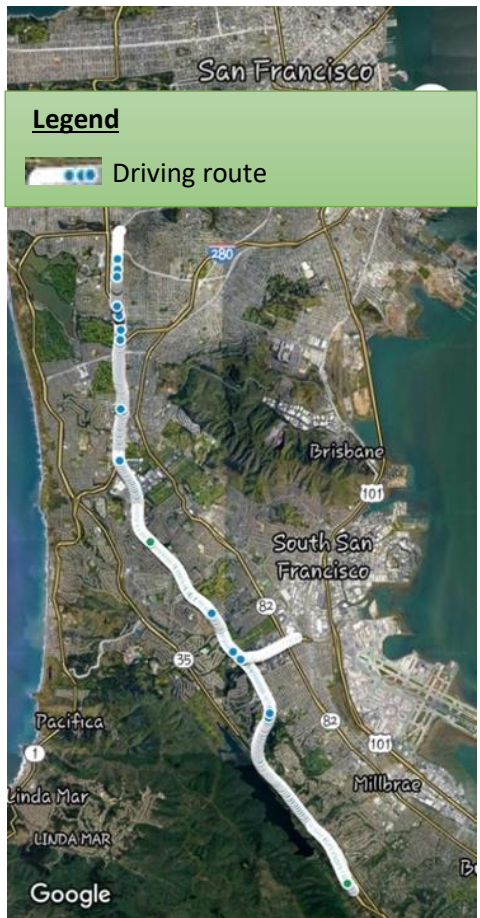

**Figure 1 Roadway segments on US-280 Highway in California.**

Along with these sensor data, the Comma2k19 dataset contains measurements from GNSS and Control Area Network (CAN) (see **Table 1** and **Table 2**). A u-blox M8 GNSS module, which can track GNSS, was used for data collection, which has a horizontal position accuracy of 2.5m. Global Positioning System (GPS) and Global Orbiting Navigational Satellite System (GLONASS) signals were used for position measurements. Further, an open-access GNSS processing library Laika was used to decrease the positioning error. It reduced the positioning error by 40%. The relevant IMU data to this study is the acceleration of the vehicle (as shown in **Table 3**).

Similarly, the relevant CAN data is the vehicle's speed and steering wheel angle data (see **Table 2**). The GNSS dataset contains live and raw GNSS data from u-blox and Qcom. Each live section includes latitude, longitude, speed, utc_timestamp, altitude, and bearing angle data. This dataset contains dense and diverse driving data which can be used to train recurrent neural network model and predict the desired variable. This dataset used for the analysis contains 7200 GNSS observations, 35726 CAN observations, and 72148 IMU observations.

**Table 1. Sample GNSS data from the comma2k19 dataset**

| GNSS time (s) | Latitude (deg) | Longitude (deg) | Speed (m/s) |
|---|---|---|---|
| 238867.5 | 37.63443 | -122.414 | 26.790 |
| 238867.6 | 37.63444 | -122.414 | 26.838 |
| 238867.7 | 37.63444 | -122.414 | 26.724 |
| 238867.8 | 37.63444 | -122.414 | 26.858 |
| 238867.9 | 37.63444 | -122.414 | 26.793 |

**Table 2. Sample CAN data from the comma2k19 dataset**

| CAN time (s) | Car speed (m/s) | Steering angle (deg) |
|---|---|---|
| 238867.4284 | 27.0375 | 2.8 |
| 238867.4535 | 27.06042 | 2.8 |
| 238867.4761 | 27.04792 | 2.8 |
| 238867.4937 | 27.04792 | 2.8 |
| 238867.5176 | 27.05347 | 2.8 |

**Table 3. Sample IMU data from the comma2k19 dataset**

| IMU time (s) | Acceleration forward (ms$^{-2}$) | Acceleration right (ms$^{-2}$) | Acceleration down (ms$^{-2}$) |
|---|---|---|---|
| 238867.4351 | 0.183838 | -0.11644 | -10.2427 |
| 238867.4451 | -0.33813 | 0.198074 | -10.1607 |
| 238867.4551 | 0.060318 | 0.057846 | -9.83777 |
| 238867.465 | -0.05409 | -0.24471 | -9.77533 |
| 238867.475 | 0.281784 | -0.28769 | -9.38159 |

For developing the prediction-based GNSS spoofing attack detection strategy, we use latitude, longitude, and speed data from GNSS, speed and steering angle data from CAN, and forward, right and down acceleration data from IMU. **Figure 2** presents all the observations from GNSS, CAN, and IMU, which we use in this study. If a spoofer attacks the GNSS of an AV to create a sophisticated attack, the spoofer must generate a GNSS signal where the signal strength should be higher than the AV GNSS signal strength (*27*)(*28*). As the GNSS signal is spoofed, data from GNSS will be corrupted. The vehicle will be directed to the wrong destination as per the spoofed data. If we can predict the distance traveled between the current location and the immediate future location of an autonomous vehicle, then we can compare it with the distance traveled between the current location and the immediate future location when the AV will be at the future timestamp. For developing a prediction-based detection strategy, we use CAN and IMU data along with the current GNSS data assuming that current data are not spoofed. The frequency of GNSS, CAN, and IMU data is 10, 50, and 100 Hz, respectively. GNSS time is used as the reference time, and all other sensor data are synchronized to prepare the train and test dataset for the LSTM model. To obtain CAN and IMU) data at the exact time as the reference time, interpolation is performed between the two closest observations in which the GNSS observation exists.

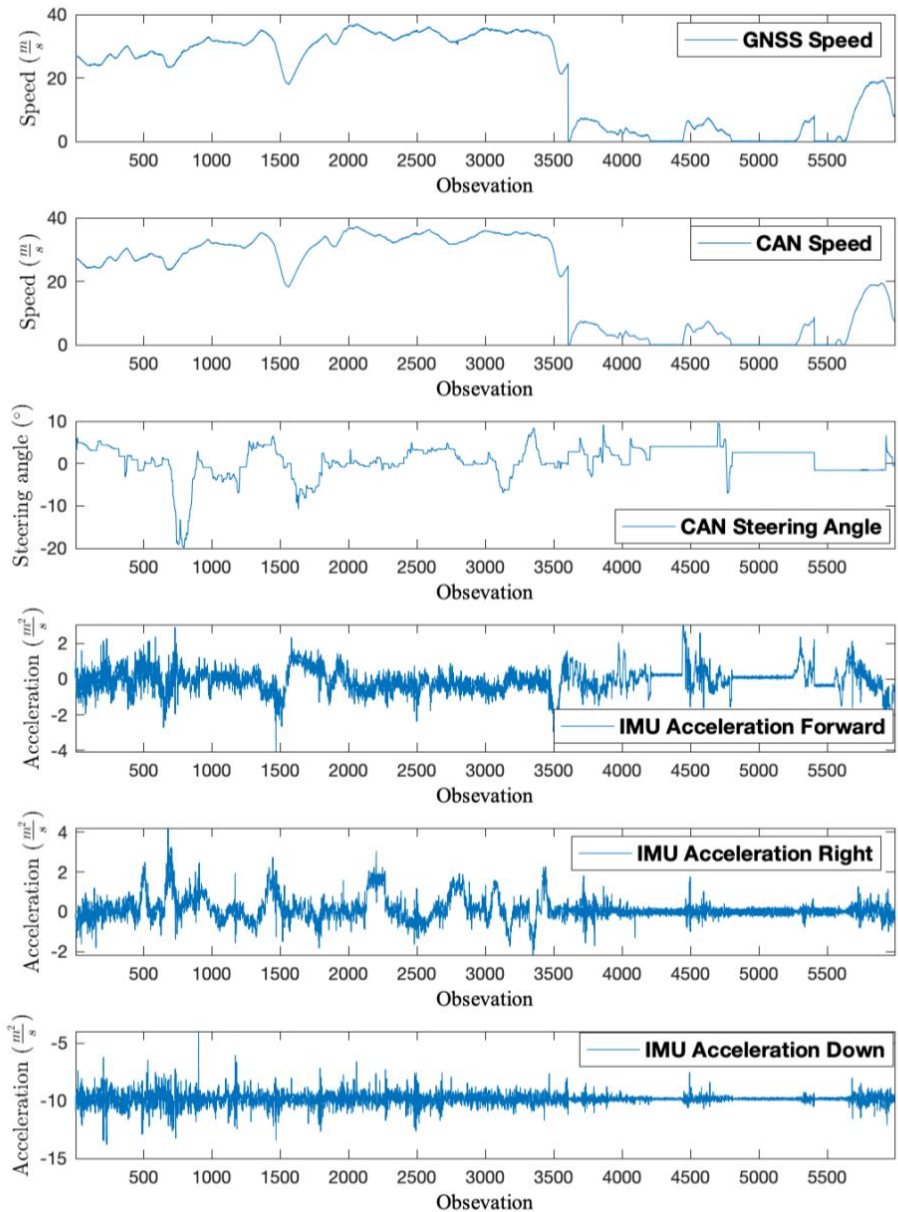

**Figure 2 Observations from GNSS, CAN and IMU**

As we predict the distance traveled between the current location and the immediate future location of an autonomous vehicle, the distance traveled in each time step from the previous time step is calculated using the latitude and longitude coordinates and the following Haversine great circle formula:(*29*)

$$d = 2r \sin^{-1}\left(\sqrt{\sin^2\left(\frac{\varphi_2 - \varphi_1}{2}\right) + \cos(\varphi_1)\cos(\varphi_2)\sin^2\left(\frac{\psi_2 - \psi_1}{2}\right)}\right) \quad (1)$$

where d is the distance between two points on the Earth's surface, r is the radius of the Earth, $\varphi_1, \varphi_2$ are latitude in radians of the current location and $\psi_2, \psi_1$ are longitudes in radians of the immediate future location and the *haversine* function is defined as,

$$hav(\theta) = \sin^2\left(\frac{\theta}{2}\right) = \frac{1 - \cos\theta}{2} \quad (2)$$

where, θ is the measurement of angle in radians. **Figure 3** presents the distance traveled between the current location and the immediate future location for each timestamp of the 5987 observations from a chunk of the comma2k19 dataset. This data along with the CAN and IMU are used to train the LSTM model.

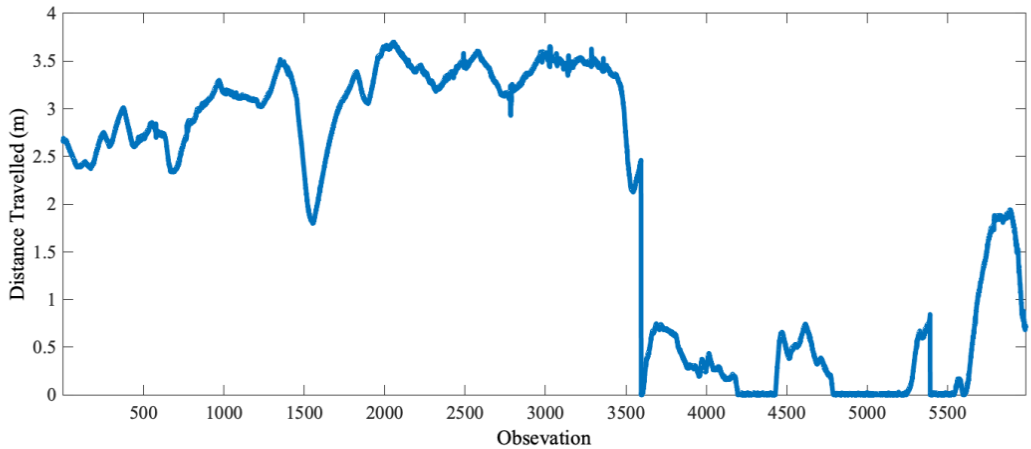

**Figure 3 Distance traveled between the current location and the immediate future location in each timestamp**

**PREDICTION MODEL DEVELOPMENT**
LSTM is a specific type of RNN, which can deal with the long-term dependencies by storing the temporal dependency of the time-series data in the memory blocks in the recurrent hidden layer. Each memory block consists of three gates: (i) input, (ii) output and (iii) forget gate. The block state is controlled by these three gates. A single LSTM block includes a sigmoid (σ) and a *tanh* activation functions. We use the LSTM model to predict the distance traveled between the current location and the immediate future location for each timestamp using the attack free CAN, IMU and GNSS sensor data (ground truth). In this paper, we have used an LSTM model, which consists of an input layer, a recurrent hidden layer with 50 neurons and an output layer. The input data to

*Dasgupta, Rahman, Islam and Chowdhury*

train the LSTM model consists of, speed and steering angle data from CAN and forward acceleration data from IMU. The output is the distance traveled between the current location and the immediate future location for each timestamp.

Before training the LSTM model, all the features are normalized between 0 and 1 and then fed into the LSTM model as input. We split the dataset into a training dataset consisting of 4500 samples, and a validation dataset consisting of 1487 samples. The hyperparameters of the LSTM model, such as the number of neurons, number of epochs, the batch size, and the learning rate play a vital role in improving prediction accuracy. A trial and error approach is used to find out optimal hyperparameters, and Adam optimizer is used as the evaluation metric to ensure there are no overfitting and underfitting issues while we train the model. The Mean Absolute Error (MAE) criterion is used as the loss function, which is defined as follows.

$$MAE = \frac{1}{N}\sum_{i=1}^{N}|y_p - y_g| \tag{3}$$

where, $N$ is the total training sample size, $y_g$ and $y_p$ are ground truth and predicted distance data respectively. The values of optimized hyperparameters are listed in **TABLE 4**. We plotted MAE on training and validation datasets to evaluate the goodness-of-fit of the LSTM model. The loss profiles using optimal hyperparameters are shown in **Figure 4** with both the training and validation datasets. A comparison of the MAE of these two datasets indicates a good fit of the prediction model with the optimal hyperparameters.

**Table 4. LSTM Model Hyperparameters.**

| Hyperparameters | Value |
|---|---|
| Number of neurons | 50 |
| Number of epochs | 100 |
| Batch size | 50 |
| Learning rate | 0.01 |

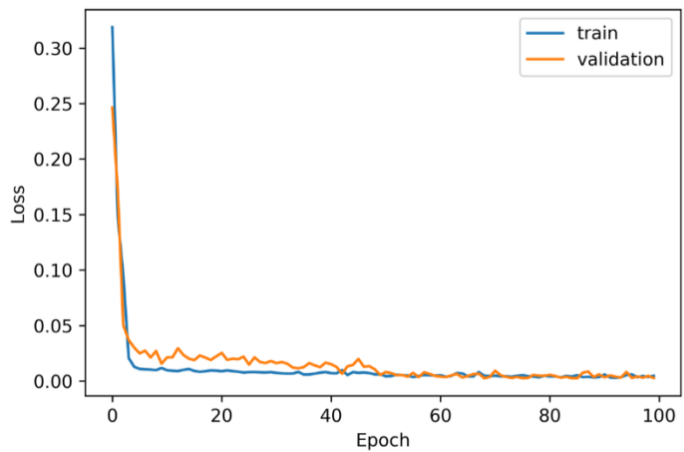

**Figure 4 Comparison of Mean Absolute Error (MAE) (loss) profiles with the optimal parameter set.**

We have tested our trained prediction model using test dataset. The RMSE is used as the evaluation metric to determine the effectiveness of the model as follows.

$$RMSE = \frac{1}{N}\sqrt{\sum_{i=1}^{N}(y_p - y_g)^2} \qquad (4)$$

Where, $N$ is the total testing sample size, $y_g$ and $y_p$ are ground truth and predicted distance data respectively. Figure 5 presents the ground truth and predicted distance traveled between the current location and the immediate future location for each timestamp. We found that the RMSE for the predicted distance traveled is 0.0242m, and the average absolute error is 0.0203 m.

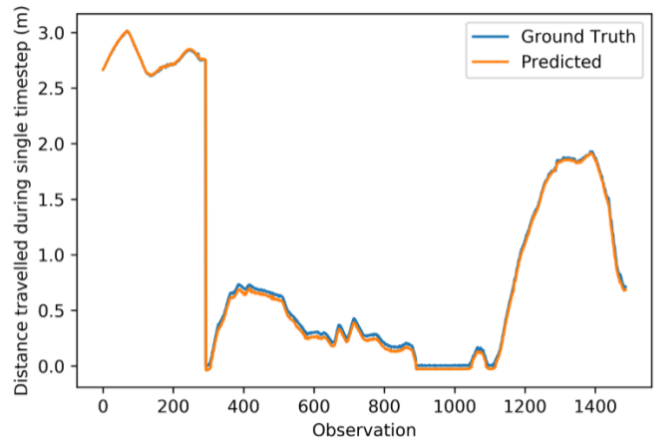

**Figure 5 Ground truth and predicted distance traveled between the current location and the immediate future location for each timestamp.**

**GNSS SPOOFING ATTACK ON AVs**
This section presents GNSS spoofed attack evaluation scenarios and attack generation, GNSS attack detection strategy, and analysis of spoofed attack detection accuracy.

**Spoofed Attack Scenario and Attack Generation**
The worst spoofing scenario for an autonomous vehicle is when an autonomous vehicle trusts the spoofed signal (*17*). When a vehicle's GNSS signal is spoofed, it will perceive a wrong position because of the spoofed data. An attacker can manipulate an autopilot's decision, which may result in rerouting the AV to a dangerous location or even can cause a fatal accident. In (*17*), the authors have discussed different types of spoofing attacks. In this study, a spoofing attack scenario is created based on the ground truth data from the comma2k19 dataset. During this attack, a spoofer manipulates the GNSS signal in such a way that the AV perceives that it is taking an exit from a freeway when it is actually steering forward. The spoofing scenario is shown in **Figure 6** where the green circles represent the positions of ground truth route and the red circles represent the route that the AV is perceiving after being spoofed. The vehicle is traveling on the Quentin L. Kopp Freeway as shown in **Figure 6**. However, the AV's GNSS receiver is spoofed just before a highway exit, the attacker generates fake GNSS signals, which mimic that the AV is taking an exit. For the spoofed data generation, the spoofed route is created using the Google Map. To make the spoofed route realistic, the positions are added using the "Add driving route" function of the Google Map, which automatically makes sure that the created route represents a feasible route. The latitude and longitude data of the spoofed route are exported as KML format.

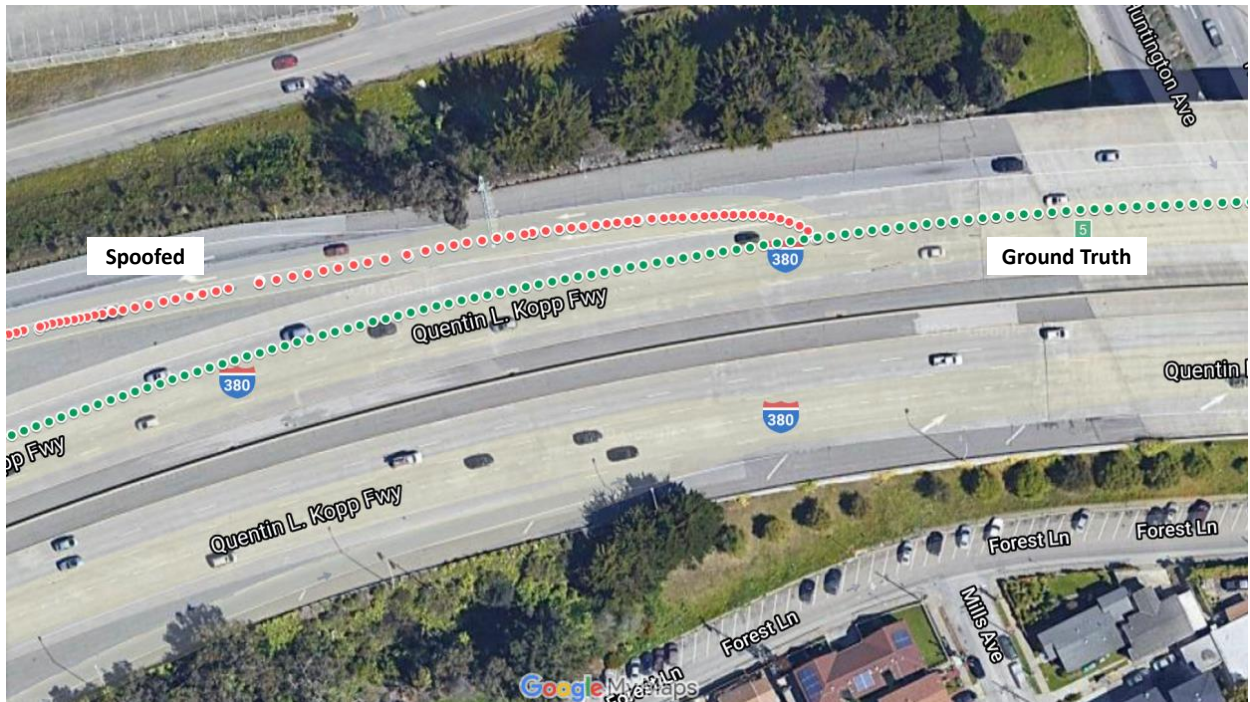

**Figure 6. Spoofed attack scenario.**

**GNSS Attack Detection Strategy**
A spoofing attack is detected by comparing the LSTM predicted distance traveled by an AV and the distance traveled based on the GNSS data between the current location and the immediate future location for each timestamp. At any time, *t*, the model will predict a distance a vehicle will travel by the *(t+1)* time step based on the current GNSS, CAN and IMU data at time step *t*. As the frequency of the GNSS receiver module is fixed, the time step duration is fixed. An error threshold is established by adding the GNSS positioning error and the LSTM prediction error. At *(t+1)* time step, the distance traveled by the vehicle is calculated using latitude and longitude at *t* and *(t+1)* time steps. If the calculated distance is larger than the spoofed attack detection threshold ($\gamma$), then our prediction-based detection model will detect it as a GNSS spoofing attack.

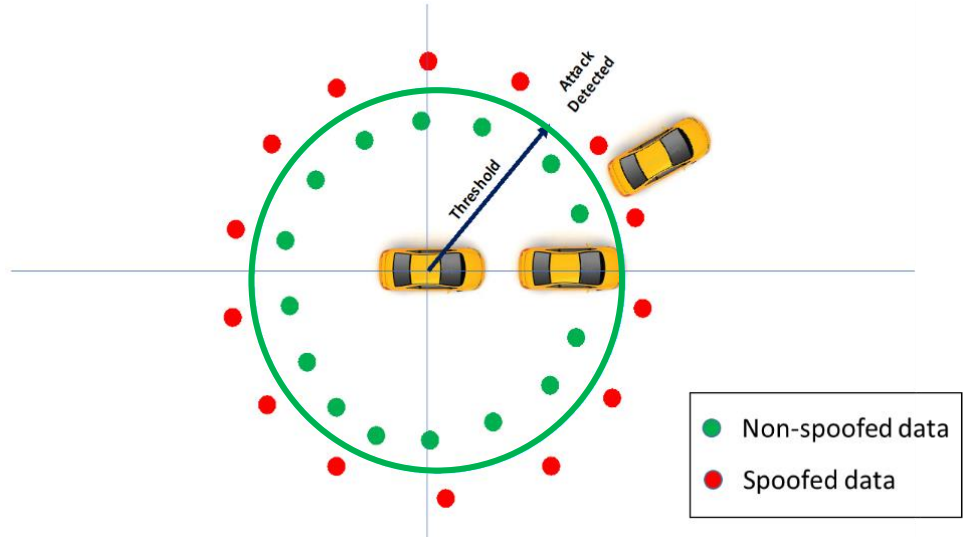

**Figure 7 GNSS attack detection strategy.**

As presented in **Figure 7**, the center of the green circle coincides with the current position of the vehicle and radius is equal to the error threshold. The error threshold is the sum of the prediction error of the LSTM model and GNSS positioning error. If the next position data of an AV from the GNSS lies inside the circle (solid green circles) then the vehicle is not under attack, but if the next position lies outside the green circle (solid red circles) then our strategy can detect it as an attack. **Figure 8** presents the histogram of the absolute prediction error for predicting distance traveled between the current AV location and the immediate future location for all observations from the test dataset. The maximum, minimum and average absolute errors are 0.0650m, 0.000046m and 0.0203m, respectively. In order to calculate the error threshold, we have used maximum absolute error as the prediction error. As the GNSS receiver module (u-blox M8) is used for GNSS data collection, the GNSS positioning error is 2.5m according to reference (*30*). By considering a 40% improvement in overall positioning error using Laika, the GNSS position error comes down to 1.5m (*26*). Thus, according to our detection strategy, the error threshold is 1.5650m ($\gamma = $ 1.5650m) for detecting a spoofing attack on an AV.

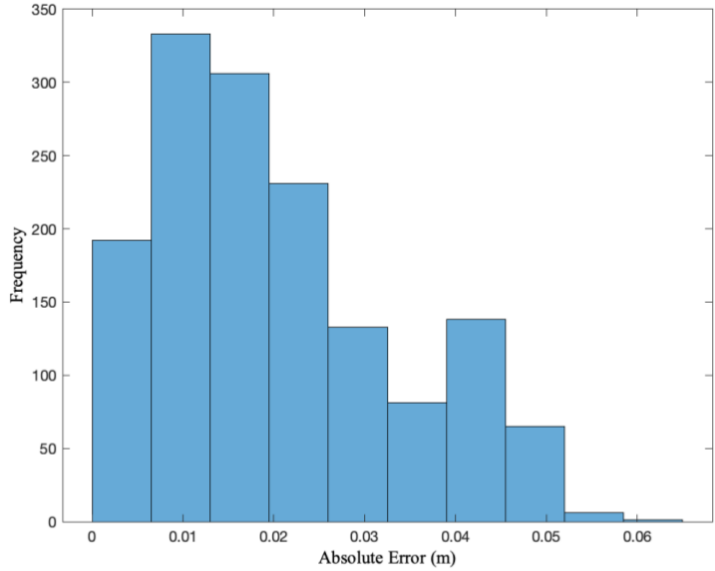

**Figure 8 Histogram of absolute prediction error.**

**Analysis of Spoofed Attack Detection Strategy**
The spoofing attack detection strategy presented in the previous section is evaluated against the scenario described in the "GNSS Attack Detection Strategy" section. **Figure 9(a)** indicates that from the point where the spoofing attack initiates, the distance between ground truth and spoofed location started to increases substantially and the point of the attack is evident from **Figure 9(a)**. **Figure 9(b)** presents the difference in distance traveled between the ground truth and spoofed location at each time step. As soon as the vehicle's GNSS receiver is compromised, the difference between the predicted distance traveled and the distance calculated based on the spoofed GNSS signal is larger than the attack detection threshold (i.e., 1.5650m). Thus, our attack detection strategy detects an attack as soon as the attack detection threshold error is exceeded. We have also calculated the computational latency for our prediction-based GNSS attack detection strategy, and the computational latency is 5 ms, which is significantly less than the real-time computational latency requirement, i.e., 100 ms (the frequency of GNSS data, 10Hz). These demonstrate that our model can successfully detect spoofing attacks.

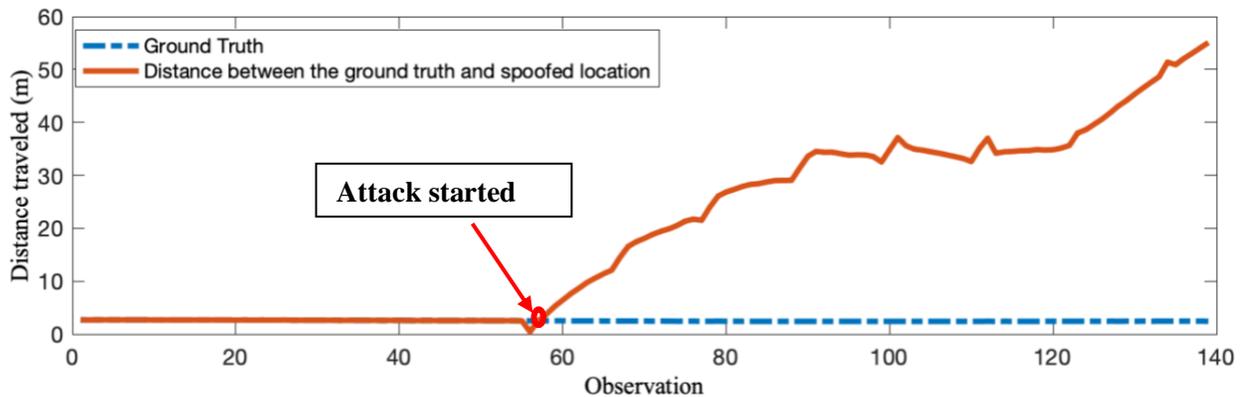

9(a)

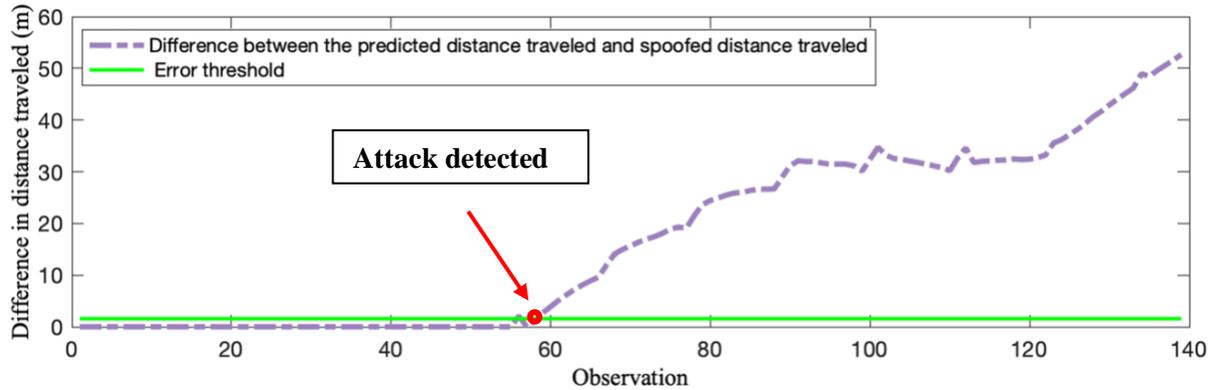

9(b)

**Figure 9 Evaluation of GNSS attack detection.**

**CONCLUSIONS**

The GNSS spoofing attack on an autonomous vehicle is one of the more sophisticated type of attacks that is difficult to detect as an attacker can mimic the GNSS signal and transmit inaccurate coordinates to an AV. Thus, spoofing attacks can compromise vehicle control that is dependent on PNT services, which can lead to serious harm to the attacked AV as well as other road users. In this study, we have developed a prediction-based spoofing attack detection strategy. We have used LSTM to predict the distance traveled between two consecutive locations of an autonomous vehicle. Based on the predicted distance traveled between the current location and the immediate future location of an autonomous vehicle, an error threshold is established using the positioning error of the GNSS device and prediction error (i.e., maximum absolute error) related to distance traveled between the current location and the immediate future location. We created an attack dataset involving AVs being attacked through GNSS spoofing to assess the efficacy of our prediction-based detection strategy. Our analysis reveals that the prediction-based spoofed attack detection strategy can successfully detect GNSS spoofing attacks in real-time. This study does not focus on the countermeasures once a GNSS spoofing attack is detected. Many existing studies focus on countermeasures to GNSS spoofing attacks on AVs.

**AUTHOR CONTRIBUTIONS**

The authors confirm contribution to the paper as follows: study conception and design: S. Dasgupta, M. Rahman, M. Islam, and M. Chowdhury; data collection: S. Dasgupta and M. Rahman; interpretation of results: S. Dasgupta, M. Rahman, M. Islam, and M. Chowdhury; draft manuscript preparation: S. Dasgupta, M. Rahman, M. Islam, and M. Chowdhury. All authors reviewed the results and approved the final version of the manuscript.